\documentclass[conference]{IEEEtran}
\IEEEoverridecommandlockouts
\usepackage{cite}
\usepackage{amsmath,amssymb,amsfonts}
\usepackage{algorithmic}
\usepackage{graphicx}
\usepackage{textcomp}
\usepackage{xcolor}
\usepackage{subfigure}
\usepackage{hyperref}
\usepackage{dblfloatfix}
\def\BibTeX{{\rm B\kern-.05em{\sc i\kern-.025em b}\kern-.08em
    T\kern-.1667em\lower.7ex\hbox{E}\kern-.125emX}}
\begin{document}

\title{Hand-Object Contact Detection using Grasp Quality Metrics}

\author{
    \IEEEauthorblockN{Thanh Vinh Nguyen}
    \IEEEauthorblockA {
        \textit{Deakin University}\\
        Melbourne, Australia \\
        0009-0003-4157-2988
    }
    \and
    \IEEEauthorblockN{Akansel Cosgun}
    \IEEEauthorblockA{
        \textit{Deakin University}\\
        Melbourne, Australia \\
        0000-0003-4203-6477
    }
}

    \maketitle

\begin{abstract}
    We propose a novel hand-object contact detection system based on grasp quality metrics extracted from object and hand poses, and evaluated its performance using the DexYCB dataset. Our evaluation demonstrated the system's high accuracy (approaching 90\%). Future work will focus on a real-time implementation using vision-based estimation, and integrating it to a robot-to-human handover system.
\end{abstract}

\section{Introduction}

Robot-to-human object handover is an active field of research, focusing on enhancing robotic assistants to interact with humans at a near-human level. A key aspect of robot-to-human handovers is determining when the human makes contact with the object and is ready to pick it up. State-of-the-art techniques on contact detection rely on physical interactions, such as force or contact sensing \cite{handover}, which often require costly sensors \cite{chan}. Furthermore, if the human is merely touching the object in a non-grasping pose, a force-based approach might result in the object being released prematurely, causing the handover to fail. In this paper, we propose an object grasping detection system based on grasp quality metrics. Our approach requires the pose estimation of the human's hand, as well as the object. We evaluate the performance of our approach using the DexYCB dataset, presenting its accuracy and proposing methods to interpret quality metrics for detecting hand-object contact.

\section{Approach}

\begin{figure}[t!]
    \centerline{
        \subfigure[Image from DexYCB]{\includegraphics[width=0.45\linewidth]{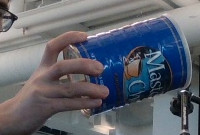}}
        \hfil
        \subfigure[Reconstructed in GraspIt!]{\includegraphics[width=0.55\linewidth]{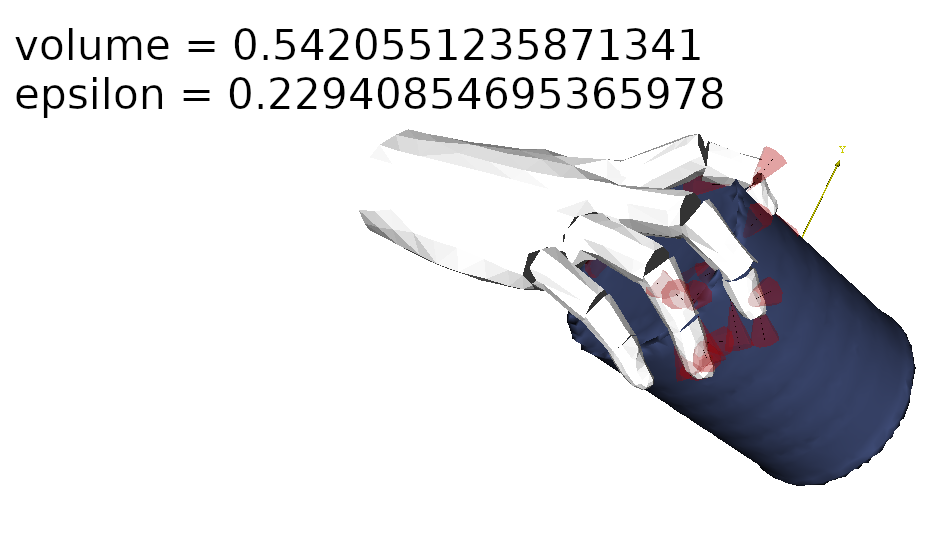}}
    }
    \centerline{}
    \caption{We detect if a hand is in contact with the object using the grasp quality metrics. We use the ground truth data from DexYCB dataset (a), reconstruct the scene in GraspIt! simulator, and extract the grasp quality metrics (b). We consider the hand in contact with the object if one of the two grasp metrics are non-zero. Our approach can also be used to determine if the hand is in a graspable pose to grasp the object.}
\end{figure}

\begin{figure}[b!]
    \centerline{\includegraphics[width=0.4\linewidth]{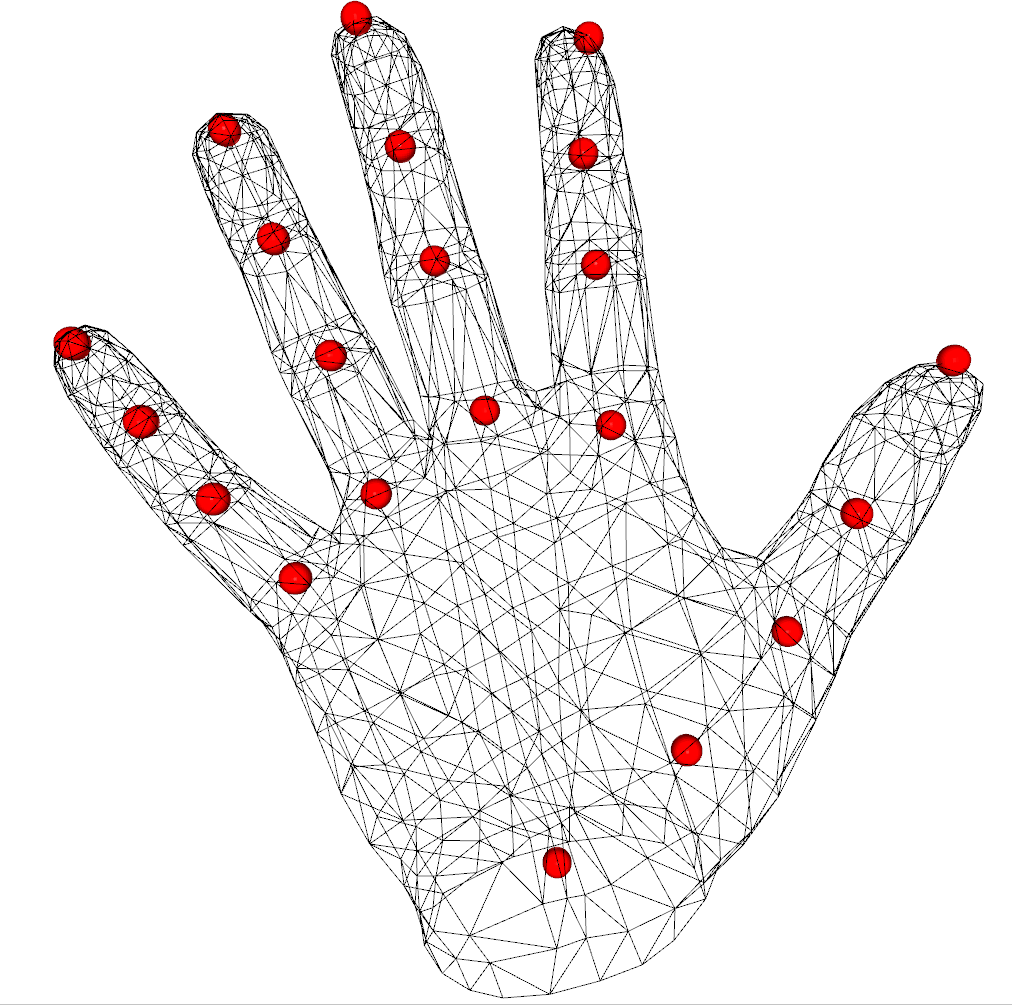}}
    \caption{Hand 3D mesh generated by the MANO model. Joints' positions are shown as dots.}
    \label{fig:joints}
\end{figure}

\begin{figure*}[ht!]
    \centerline{\includegraphics[width=0.85\linewidth]{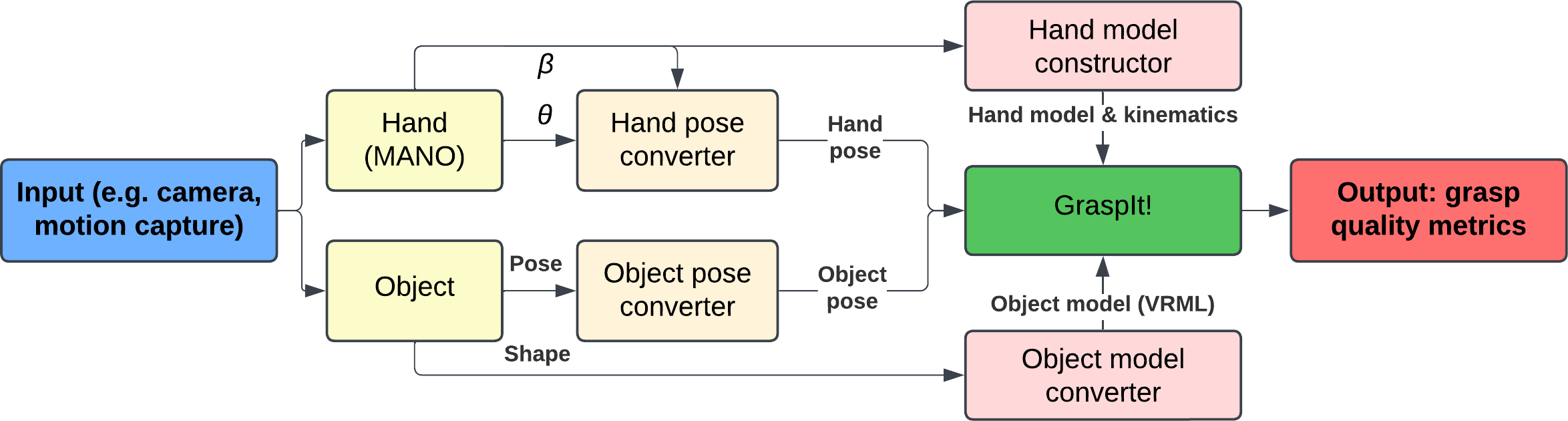}} % TODO: redraw
    \caption{The system diagram for our hand-object contact detection.}
    \label{fig:architecture}
\end{figure*}

Fig.~\ref{fig:architecture} illustrates the architecture of the system, which evaluates a human's grasp of an object by reconstructing the interaction scene in the GraspIt! simulator \cite{graspit}. For hand reconstruction, the system employs the MANO hand model \cite{mano}, which allows a human hand surface to be represented as the shape parameter vector $\beta\in\mathbb{R}^{10}$ and pose parameter vector $\theta\in\mathbb{R}^N$, the latter of which can have varying sizes through principal component analysis (PCA). The $\beta$ vector is used to generate a flat hand 3D mesh for each human, thus accounting for differences in hand shapes, which is then converted into a joint position-based kinematic model for use by the simulator. This model is then actuated using a hand pose converter routine, which computes the hand's 21 joints' coordinates (Fig.~\ref{fig:joints}) in an input frame, given the hand's $\beta$ parameters and the $\theta$ parameters captured from the frame, and computes the joint position values to provide to the simulator.

A similar scheme is implemented to reconstruct the objects in the simulated space, where the objects' 3D models are captured and converted to the Virtual Reality Modelling Language (VRML)-based format used by GraspIt! beforehand, and their poses are then set for each frame.

The GraspIt! simulator has built-in capabilities for the computation of Ferrari-Canny grasp quality measures $\epsilon$ and $v$, indicating the worst-case and average-case grasp stability respectively \cite{qm}. By reconstructing the object interaction in the simulator, it can compute the quality of a given grasping configuration, which may be used to determine whether a human has a stable grasp of an object.

\begin{table*}[h!]
    \caption{Contact detection accuracy per object. TP: True Positive, TN: True Negative, FP: False Positive, FN: False Negative.}
    \begin{center}
        \begin{tabular}{|l|c|c|c|c|c|c|}
            \hline
            \textbf{Object (YCB name)} & \textbf{Frames} & \textbf{TP (\%)} & \textbf{TN (\%)} & \textbf{FP (\%)} & \textbf{FN (\%)} & \textbf{Accuracy (\%)} \\
            \hline
            002\_master\_chef\_can & 1265 & 93.9 & 80.4 & 19.6 & 6.1 & 89.4 \\
            \hline
            003\_cracker\_box & 1137 & 91.5 & 77.0 & 23.0 & 8.5 & 86.2 \\
            \hline
            004\_sugar\_box & 1125 & 94.9 & 84.4 & 15.6 & 5.1 & 91.6 \\
            \hline
            005\_tomato\_soup\_can & 1188 & 96.3 & 83.3 & 16.7 & 3.7 & 92.0 \\
            \hline
            006\_mustard\_bottle & 1094 & 94.8 & 79.9 & 20.1 & 5.2 & 89.9 \\
            \hline
            007\_tuna\_fish\_can & 1213 & 89.6 & 87.7 & 12.3 & 10.4 & 89.0 \\
            \hline
            008\_pudding\_box & 1089 & 96.1 & 87.4 & 12.6 & 3.9 & 93.2 \\
            \hline
            009\_gelatin\_box & 1177 & 94.7 & 87.8 & 12.2 & 5.3 & 92.6 \\
            \hline
            010\_potted\_meat\_can & 1091 & 96.1 & 80.6 & 19.4 & 3.9 & 91.1 \\
            \hline
            011\_banana & 1002 & 87.3 & 92.5 & 7.5 & 12.7 & 89.0 \\
            \hline
            019\_pitcher\_base & 1235 & 92.6 & 81.3 & 18.7 & 7.4 & 88.8 \\
            \hline
            021\_bleach\_cleanser & 1017 & 94.6 & 74.1 & 25.9 & 5.4 & 86.9 \\
            \hline
            024\_bowl & 1232 & 84.9 & 82.4 & 17.6 & 15.1 & 83.9 \\
            \hline
            025\_mug & 1041 & 95.5 & 87.8 & 12.2 & 4.5 & 93.0 \\
            \hline
            035\_power\_drill & 1011 & 95.1 & 70.6 & 29.4 & 4.9 & 85.6 \\
            \hline
            036\_wood\_block & 1163 & 93.2 & 79.3 & 20.7 & 6.8 & 88.3 \\
            \hline
            037\_scissors & 1085 & 90.1 & 90.7 & 9.3 & 9.9 & 90.3 \\
            \hline
            040\_large\_marker & 1159 & 83.2 & 97.0 & 3.0 & 16.8 & 87.6 \\
            \hline
            052\_extra\_large\_clamp & 1142 & 87.8 & 89.2 & 10.8 & 12.2 & 88.3 \\
            \hline
            061\_foam\_brick & 1193 & 87.9 & 94.9 & 5.1 & 12.1 & 90.4 \\
            \hline
            \textbf{Overall} & \textbf{22659} & \textbf{91.8} & \textbf{84.3} & \textbf{15.7} & \textbf{8.0} & \textbf{89.3} \\
            \hline
        \end{tabular}
    \end{center}
    \label{tab:acc_per_object}
\end{table*}

\begin{table*}[ht!]
    \caption{Contact detection accuracy per human subject. TP: True Positive, TN: True Negative, FP: False Positive, FN: False Negative.}
    \begin{center}
        \begin{tabular}{|c|c|c|c|c|c|c|}
            \hline
            \textbf{Subject} & \textbf{Frames} & \textbf{TP (\%)} & \textbf{TN (\%)} & \textbf{FP (\%)} & \textbf{FN (\%)} & \textbf{Accuracy (\%)} \\
            \hline
            1 & 3463 & 90.5 & 89.0 & 11.0 & 9.5 & 89.9 \\
            \hline
            2 & 3542 & 92.9 & 85.3 & 14.7 & 7.1 & 90.3 \\
            \hline
            3 & 3158 & 92.2 & 85.0 & 15.0 & 7.8 & 90.6 \\
            \hline
            4 & 2860 & 94.9 & 85.0 & 15.0 & 5.1 & 90.6 \\
            \hline
            6 & 3074 & 94.8 & 83.8 & 16.2 & 5.2 & 91.1 \\
            \hline
            8 & 2947 & 87.0 & 82.4 & 17.6 & 13.0 & 85.1 \\
            \hline
            9 & 3615 & 91.5 & 78.2 & 21.8 & 8.5 & 87.8 \\
            \hline
            \textbf{Overall} & \textbf{22659} & \textbf{91.8} & \textbf{84.3} & \textbf{15.7} & \textbf{8.0} & \textbf{89.3} \\
            \hline
        \end{tabular}
    \end{center}
    \label{tab:acc_per_subject}
\end{table*}

The system was implemented using the Python programming language, with the GraspIt! simulator being operated through its Robot Operating System (ROS) interface. Our implementation utilises the Manopth library, originally developed for \cite{manopth}, to integrate the MANO hand model with the system. Hand models were initially converted to Unified Robot Description Format (URDF) descriptions, which were then converted to robot descriptions used by the simulator.

\section{Evaluation}

Our approach was evaluated using the DexYCB dataset \cite{dexycb}, which consists of annotated images and pose data of human-object interactions performed by 10 human subjects. The dataset is based on the Yale-CMU-Berkeley (YCB) object set \cite{ycb}, which consists of various daily-life objects having different shapes and sizes and is used extensively in robotic grasping research. To simplify our evaluation, we use the ground truth hand and object poses in the global reference frame. Hand pose parameters are provided by the dataset as vectors of size $N=48$. For each frame, only the hand and the grasped object were reconstructed in the Graspit simulator. Due to technical issues with hand model generation, our evaluation only included right-handed grasps of 7 human subjects, amounting to 351 grasping sequences and 22659 data points.

When no contact between the hand and the object is detected, the simulator reports the grasp quality measures as $\epsilon=-1.0$ and $v=0.0$, and when there is contact, we observed that at least one of these metrics will be a different value. We therefore define the system's criteria for hand-object contact detection as either the $\epsilon$ or $v$ measure being greater than zero. The system's detections were compared to the ground truth data, which was manually annotated to identify frames where the subjects appeared to achieve a stable grasp. Statistical analysis of the obtained grasp quality metrics, accuracy measures, and processing times was conducted to gain further insights into the system. The evaluation results are detailed in the following section.

\subsection{Detection accuracy}

\begin{figure}[h!]
    \centerline{
        \subfigure[024\_bowl]{\includegraphics[width=0.3\linewidth]{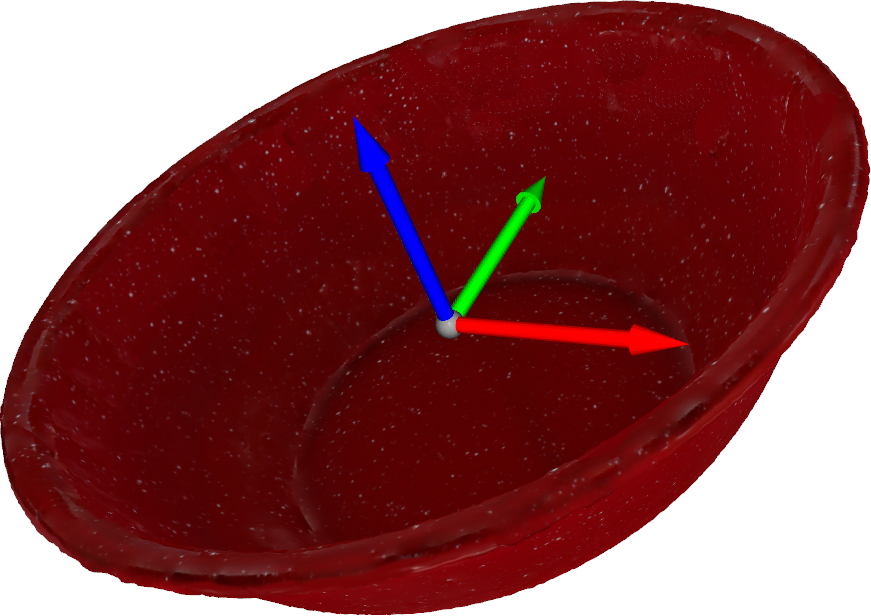}\label{fig:024_bowl}}
        \hfil
        \subfigure[008\_pudding\_box]{\includegraphics[width=0.3\linewidth]{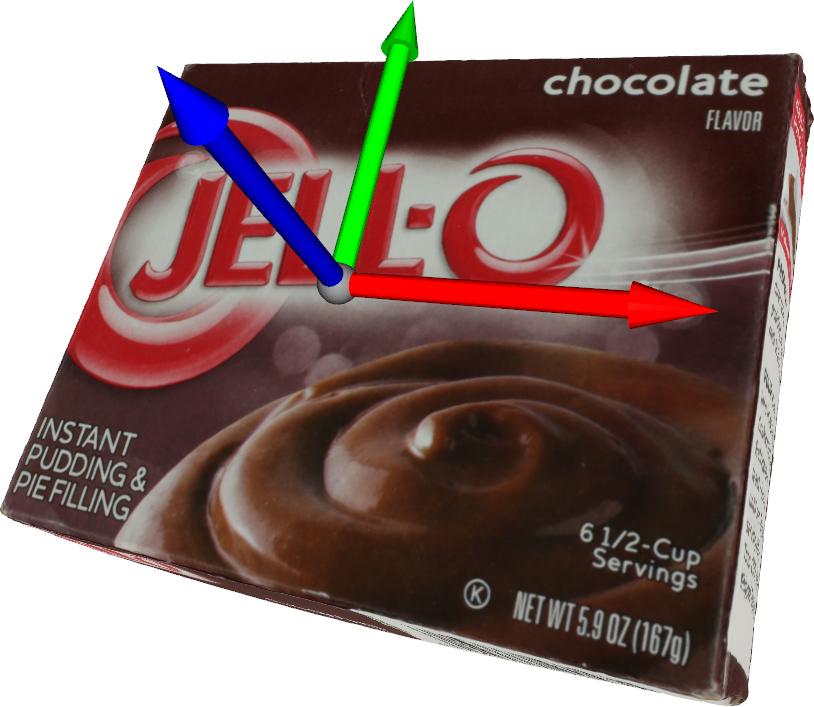}\label{fig:008_pudding_box}}
    }
    \caption{3D models of the (a) least and (b) most accurately detected objects. A $5\times5\times5$ cm coordinate frame is included in each visualisation for scale.}
\end{figure}

Table~\ref{tab:acc_per_object} reports the accuracy of this contact detection method with respect to objects being grasped by the subjects. Overall, our system correctly detected hand object contacts in 89.3\% of our evaluated frames, with false positive and false negative rates of 15.7\% and 8.0\% respectively. The detection accuracy varied with different objects, ranging from 83.9\% to 93.2\%. This can be attributed to the shapes and sizes of the grasped objects. For example, the object least accurately detected, 024\_bowl (Fig.~\ref{fig:024_bowl}), can be either pinched on both sides of its slim wall (thus potentially leading to the fingers being closer or even in contact with each other) or grasped from underneath (thus potentially leading to a less stable grasp). Meanwhile, the grasped object having most accurate detections, 008\_pudding\_box (Fig.~\ref{fig:008_pudding_box}), is relatively small and can be stably grasped on its sides.

\begin{figure}[h!]
    \centerline{\includegraphics[width=0.85\linewidth]{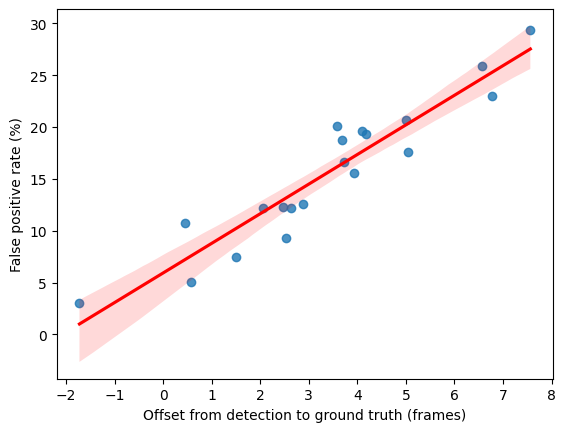}}
    \caption{The average offset between the first contact detection frame and the ground truth for each object with respect to the corresponding false positive rate. The regression line (in red) is included to highlight the correlation between these attributes.}
    \label{fig:avgoff_object}
\end{figure}

False positive rates varied significantly across objects, ranging from 3.0\% to 29.4\%. We hypothesise that the variation is primarily due to differences in the complexity of grasping hand poses for each object. Specifically, objects that take more time to grasp may be detected as in graspable contact with the hand before the stable grasp is confirmed in the ground truth, resulting in false positive frames during this transition phase. This is demonstrated in Fig.~\ref{fig:avgoff_object}, in which a correlation between the average offset from the system's detection to the ground truth and increasing false positive rates can be seen. Meanwhile, false negative rates have a lower range of 3.9\% to 16.8\% across objects. Notably, the 040\_large\_marker object has both the lowest false positive rate and the highest false negative rate. This may be caused by the object's small size and shape, which both allows human subjects to quickly grasp it, while being more susceptible to errors in the data acquisition and conversion processes and result in inaccurate reconstructions in the simulator.

Table~\ref{tab:acc_per_subject} shows our system's accuracy across the evaluated human subjects. As anticipated, our system yields similar accuracy across different subjects, with the rate ranging from 85.1\% to 91.1\%. However, evaluations with subjects 8 and 9 showed a slight decrease in accuracy, which may stem from errors in our manual ground truth annotation process.

\subsection{Grasp quality metrics}

\begin{figure}[t!]
    \centerline{
        \subfigure[$v$]{\includegraphics[width=0.45\linewidth]{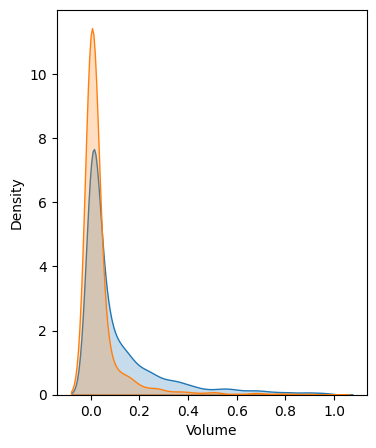}\label{fig:volume_dist}}
        \hfil
        \subfigure[$\epsilon$]{\includegraphics[width=0.45\linewidth]{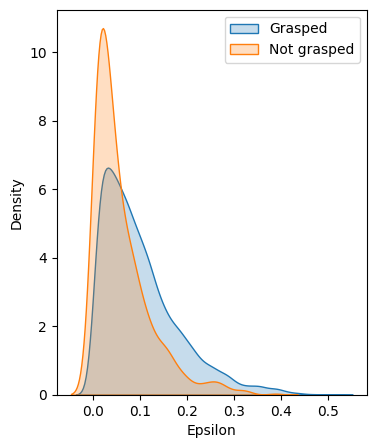}\label{fig:epsilon_dist}}
    }
    \caption{Distribution plots of grasp quality measures for grasped and non-grasped ground truths. Only values deviating from the non-contacting values $v=0.0$ and $\epsilon=-1.0$ are considered.}
    \label{fig:metrics}
\end{figure}

Fig.~\ref{fig:metrics} illustrates the distributions of the Ferrari-Canny grasp quality measures retrieved from GraspIt! with respect to the ground truth. It can be seen that both measures are centred at the same point (approx. 0.03 and 0.05 for $v$ and $\epsilon$ respectively) and have roughly similar distributions regardless of ground truth. This will complicate the interpretation of these quality measures, and subsequent implementation of a simple threshold-based object grasping detection method. It can also be seen that higher measure values are slightly more common in frames where the human subject is successfully grasping the objects, which demonstrates the relations between higher quality metrics and more stable grasping poses.

\begin{figure}[h!]
    \centerline{\includegraphics[width=0.9\linewidth]{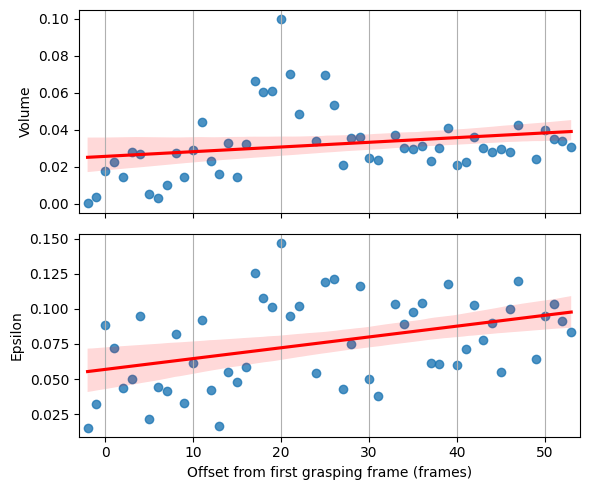}}
    \caption{Grasp quality measures $v$ (top) and $\epsilon$ (bottom) for a typical grasp sequence from the DexYCB dataset. Frames without hand-object contact as detected by the system are omitted. The X-axis represents the frame number's offset from the first grasping frame in the ground truth.}
    \label{fig:seq_metrics}
\end{figure}

Another possible approach to interpreting the grasp quality metrics is to observe their changes over time. The metrics' time series plots in a typical grasp sequence are shown in Fig.~\ref{fig:seq_metrics}, which illustrates a slight increase in both measures' values over time. A sudden increase in both $v$ and $\epsilon$ can also be seen when the sequence's ground truth state transitions from non-grasping to grasping. These observations may be taken into account for interpreting these metrics and improving the accuracy of hand object grasping in the future.

\section{Future Work}

Our future work will focus on adapting our hand-object contact detection system for operating on real-world inputs in real time. Numerous studies, such as \cite{3dhand} and \cite{wilor}, have explored hand shape parameter estimation, particularly using a single RGB image as input. Since hand shape parameters are unique to each individual, shape estimation could be performed once per subject to enhance the system's performance. In these prior works, hand shape estimation is usually integrated with pose estimation, making it possible to use a single model to reconstruct human hands in the simulator; however, more lightweight models such as MediaPipe \cite{mphands} could be employed to achieve real-time hand pose extraction and reconstruction.

Vision-based object localization and pose estimation is an extensively studied field, with various published works such as \cite{fs6d}, which could be adapted to enable object reconstruction within the system. In the context of robot-to-human handovers, the pose information from the robotic manipulator can assist with this task. There have also been efforts to estimate both hand and object poses simultaneously \cite{honnotate}, which could be implemented to simplify the scene reconstruction pipeline. Multiple camera angles could also be incorporated to address visual occlusion and improve the system's reconstruction accuracy.

To assess the system's real-world performance, we plan to integrate it with a robot-to-human handover system, enabling the comparison of its performance against established tactile-based contact detection methods.

\section{Conclusion}

In this paper, we described a hand-object contact and grasping detection approach based on grasp quality metrics. We evaluated our implementation of the system using the DexYCB dataset, which demonstrated a reasonably high detection accuracy of 89.3\%. While our evaluation relied on idealized conditions with ground truth poses for the human hand and object, future work will focus on real-time implementation using computer vision for pose estimation. We plan to integrate our vision-based contact detection approach into a robot-to-human handover system, where object release is triggered by contact detection, and compare its performance with force-based contact detection methods.

% \vspace{12pt}
% \color{red}
% IEEE conference templates contain guidance text for composing and formatting conference papers. Please ensure that all template text is removed from your conference paper prior to submission to the conference. Failure to remove the template text from your paper may result in your paper not being published.

\end{document}